\documentclass[runningheads]{llncs}

\usepackage{eccv}

\usepackage{eccvabbrv}

\usepackage{graphicx}
\usepackage{booktabs}
\usepackage{adjustbox}
\usepackage{bm}
\usepackage{verbatim}
\usepackage{xcolor,colortbl}
\usepackage{subcaption}
\usepackage{gensymb}
\usepackage{lipsum}
\usepackage{amssymb}
 \usepackage{comment}
\usepackage{arydshln}
\usepackage{mwe}
\usepackage{capt-of,etoolbox}

\usepackage[accsupp]{axessibility}  

\usepackage{hyperref}

\usepackage{orcidlink}

\begin{document}
\newcommand{\ours}{RoGUENeRF}

\newcommand{\cellbest}{\cellcolor{red!20}}
\newcommand{\cellsecond}{\cellcolor{orange!20}}
\newcommand{\boxtick}{\makebox[0pt][l]{$\square$}\raisebox{.15ex}{\hspace{0.1em}$\checkmark$}}

\title{RoGUENeRF: A Robust Geometry-Consistent Universal Enhancer for NeRF}

\author{Sibi Catley-Chandar\inst{1,2} \and
Richard Shaw\inst{1}\and \\
Gregory Slabaugh\inst{2}\and 
Eduardo P\'erez-Pellitero\inst{1}}

\authorrunning{S.~Catley-Chandar et al.}


\institute{Huawei Noah's Ark Lab\and 
Queen Mary University of London}

\maketitle

\begin{figure}[h!]
\centering
\includegraphics[width=1.0\textwidth]{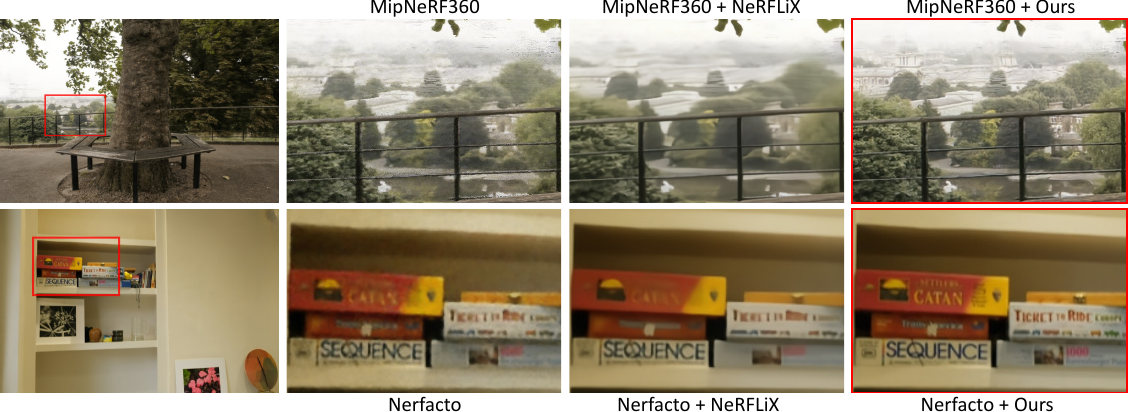}
\caption{Novel views from the MipNeRF360 dataset \cite{barron22}. \ours \ achieves noticeable qualitative improvements over state-of-the-art baselines and NeRF enhancers, especially in high-frequency regions such as trees, buildings and text.}
\label{fig:teaser-fig}
\end{figure}

\begin{abstract}

Recent advances in neural rendering have enabled highly photorealistic 3D scene reconstruction and novel view synthesis. Despite this progress, current state-of-the-art methods struggle to reconstruct high frequency detail, due to factors such as a low-frequency bias of radiance fields and inaccurate camera calibration. One approach to mitigate this issue is to enhance images post-rendering. 2D enhancers can be pre-trained to recover some detail but are agnostic to scene geometry and do not easily generalize to new distributions of image degradation. Conversely, existing 3D enhancers are able to transfer detail from nearby training images in a generalizable manner, but suffer from inaccurate camera calibration and can propagate errors from the geometry into rendered images. We propose a neural rendering enhancer, \ours, which exploits the best of both paradigms. Our method is pre-trained to learn a general enhancer while also leveraging information from nearby training images via robust 3D alignment and geometry-aware fusion. Our approach restores high-frequency textures while maintaining geometric consistency and is also robust to inaccurate camera calibration. We show that \ours~substantially enhances the rendering quality of a wide range of neural rendering baselines, e.g. improving the PSNR of MipNeRF360 by 0.63dB and Nerfacto by 1.34dB on the real world 360v2 dataset. Project page: \url{https://sib1.github.io/projects/roguenerf/}
\end{abstract}

\section{Introduction}
\label{sec:intro}

The seminal work of Mildenhall \etal~\cite{mildenhall20} introduced an effective methodology to render highly photorealistic novel views of 3D scenes by means of Neural Radiance Fields (NeRFs). Given a set of posed multi-view images, NeRFs learn complex view-dependent effects via a learnable multilayer perceptron (MLP) which models the 3D radiance field of the scene, thanks in part to the input domain parameterization (3D coordinates + 2D viewing direction) and the direct pixel-wise photometric loss. The NeRF paradigm has been very popular in recent years~\cite{xie22}, with active research in the field proposing new functionalities \cite{peng21,peng23,niemeyer20,wang22-clipnerf}, applications \cite{moreau21, mildenhall22, isik23} and also tackling some of the open challenges present in \cite{mildenhall20}. A key aspect of subsequent literature is the successive improvement of the rendering fidelity, \ie as measured by the Peak Signal-to-Noise Ratio (PSNR), producing higher quality rendered novel views \cite{barron22, barron23}.

Nonetheless, an underlying challenge of these approaches is the shape-radiance ambiguity~\cite{zhang20}, \ie training images can be explained with high accuracy by introducing inaccurate geometry, 
resulting in poor generalization outside of the training views. This is particularly problematic when the geometry around high-frequency details, such as textures, can not be resolved or disambiguated properly with the number of training views. The optimization process will generally either introduce inaccurate geometry with view-dependent radiance values overfitting each training view, \eg view-dependent floater artifacts, or else converge to a mean radiance value, which then results in blurred renderings. Similarly, inaccurate camera calibration or missing lens distortion models also lead to blurred results and thus lack of fidelity in the high-frequency spectrum, mostly due to pixel and subpixel shifts among camera views. Further, analysis by Tancik \etal~\cite{tancik20ffn} describes a low-frequency bias of the standard MLP set-up.

In addition to these issues, the practical nature of data capture and camera pose estimation can introduce further error. For pseudo-static 3D scenes, small variations in the environment can occur during capture such as changes in lighting and small movements within the scene (e.g. foliage), which violate the 3D-consistent static scene assumption of NeRFs. This can also negatively affect the performance of camera pose estimation via COLMAP \cite{schoenberger16sfm}, which although mostly reliable is not infallible \cite{raoult17, jiang23}, even when using carefully captured data.

In this work, we present \ours, a NeRF enhancer which is designed to improve the image quality of NeRF renderings while maintaining geometric consistency and being robust to inaccurate camera calibration. Firstly, we propose a novel combined 3D + 2D alignment and refinement mechanism which accurately finds correspondences between images from different camera viewpoints, even when one input is severely degraded, and can also compensate for inaccurate estimates of scene geometry and camera poses. Secondly, we propose a novel geometry-aware spatial attention module which regulates misaligned regions based on both camera distance and pixel-wise differences. Lastly, we propose a pre-training and fine-tuning strategy which learns a general geometry-consistent enhancement function that transfers well (\ie fine-tunes in under $60$ minutes per scene) to different distributions of rendering degradations. We thoroughly evaluate our proposed approach on six different NeRF baselines across real world bounded and unbounded scenes from three datasets: LLFF\cite{mildenhall19}, DTU \cite{jensen14} and 360v2 \cite{barron22}. We show consistent improvements in PSNR, SSIM and LPIPS over every baseline and qualitatively demonstrate substantial improvements in image quality over baselines and state-of-the-art NeRF enhancers.

\section{Related Work}
\label{sec:related-work}

\subsection{High Fidelity NeRF} Since the seminal work of Mildenhall \etal ~\cite{mildenhall20}, there have been several works which aim to improve the fidelity of NeRF-based models. Some methods \cite{jiang23, roessle23} incorporate additional processing layers after the NeRF model, coupled with image quality specific loss functions. Roessle \etal \cite{roessle23} proposed GANeRF, using a 2D conditional generator trained adversarially to refine the rendered output. To maintain view-consistency, a discriminator is trained end-to-end together with the underlying NeRF model, requiring computationally expensive patch-based training. Combined with the time required to train the generator, this significantly increases optimization time per scene from 15 minutes for the underlying NeRF model to 58 hours. AlignNeRF \cite{jiang23} also incorporates additional processing layers trained end-to-end with the NeRF model by introducing a shallow convolutional network coupled with an alignment aware loss and relies on the shallowness of the enhancement network to maintain view consistency. Some methods tackle the problem of NeRF super-resolution \cite{wang22-nerfsr, huang23} by using relevant patches from a HR reference frame. Other approaches attempt to improve the underlying differentiable rendering algorithm directly \cite{barron21, barron22, verbin22, xu23} to better tackle anti-aliasing effects, unbounded scenes or reflections and can better model the characteristics of the scene. A further approach is to jointly optimize camera poses together with the NeRF to reduce geometric errors from incorrect poses \cite{lin21, bian23, truong23, meuleman23, park23}.

Another popular line of work has been to address the slow training and inference time of NeRF by changing from an MLP to a faster representation, e.g.\,voxels \cite{yu21}, SDFs \cite{turki24}, MPIs \cite{tanay23, tanay24}, hash encodings \cite{mueller22, wang22-neus2, tancik23}, tensor decomposition \cite{chen22}, octrees \cite{yu21-2}, reducing training time per scene to as fast as a few seconds. ZipNeRF \cite{barron23} marries the advantages of both high-fidelity and fast-training by combining voxel-grid data structures with cone-based rendering. Dynamic scenes captured as volumetric videos also pose a significant challenge for many existing neural rendering models which otherwise perform well on static fixed scenes. Many methods model the additional time dimension \cite{pumarola20,li22, peng23, liu23, isik23, shaw23, moreau24, dhamo23} and are able to render short free viewpoint videos effectively but at the cost of lower image quality and temporal consistency. Despite huge progress in improving the fidelity of NeRFs, these models remain constrained by errors in training data and camera pose estimation leading to rendering artifacts in real world scenes. 

\subsection{NeRF Enhancers} Another approach to improving NeRF fidelity is to apply a NeRF-specific enhancer as a post-rendering step, which assumes the NeRF has already been trained and does not backpropagate gradients to the underlying NeRF model. Boosting View Synthesis \cite{rong22} aims to improve image quality by transferring colour residuals from training views to novel views inspired by concepts from classic image-based rendering methods. The use of residuals depends on a pixel perfect alignment between the rendered and ground truth training views, however this is often not the case due to inaccurate camera pose estimation \cite{jiang23} thus limiting the possibility of accurate residual transfer and requiring a hand-crafted weighting strategy to alleviate ghosting artifacts. NeRFLiX~\cite{zhou23} on the other hand eschews classic 3D models entirely and instead proposes to learn a general 2D viewpoint mixer which is trained via simulated image degradation. This is reasonably effective provided the testing domain is well represented in the simulated degradation and neighbouring images are close enough in camera pose and image content. However if the distribution of the rendering artifacts shifts from the simulated data, the performance degrades. NeRFLiX++~\cite{zhou23} significantly improves the computational efficiency of NeRFLiX while also increasing the realism of image degradations by introducing a GAN-based degradation simulator, further boosting performance. Existing NeRF enhancers either use a 2D approach to learn a general function which is agnostic to scene geometry, or they use a 3D approach which suffers from inaccurate camera calibration and can propagate errors from the geometry into image renderings. In contrast, our method is robust to errors in camera poses and maintains view-consistency while also being pre-trained to learn a general enhancement function, effectively combining the advantages of 2D and 3D approaches.

\section{Method}

\begin{figure*}[t]
\centering
\includegraphics[width=1.00\textwidth]{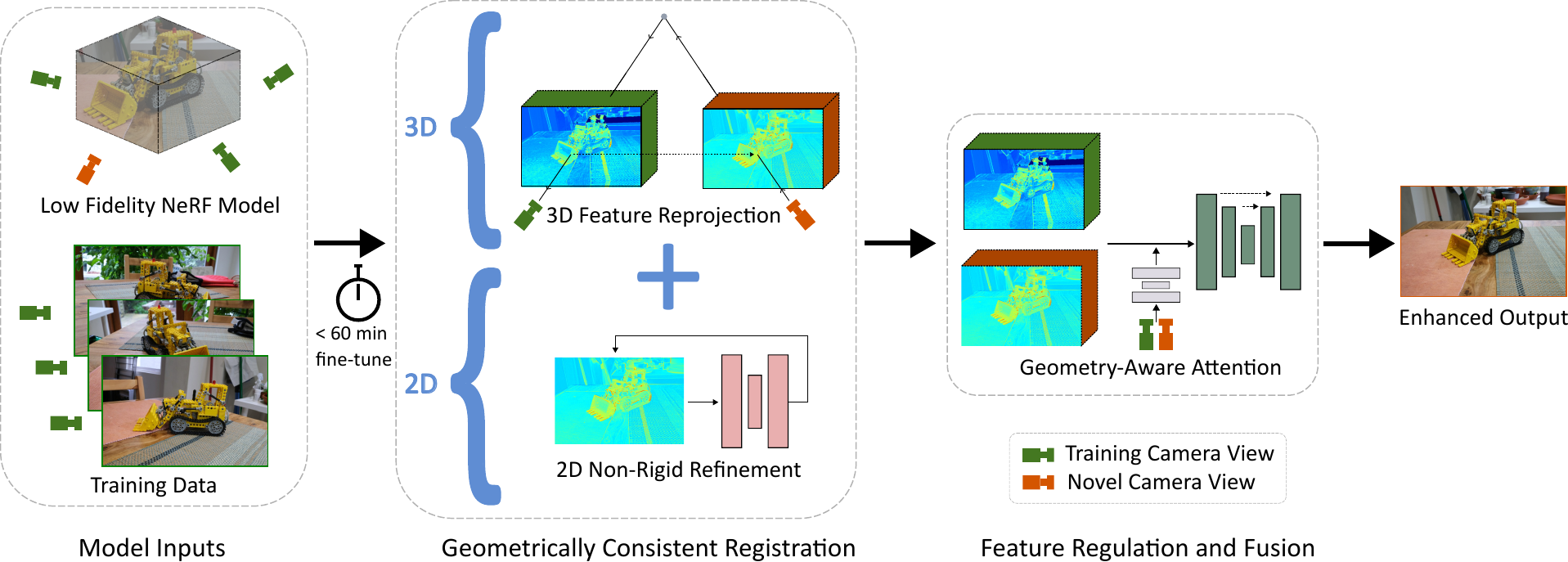}
\caption{\ours \ Overview: Given a trained NeRF model and corresponding training data, our method substantially enhances the rendering quality of the NeRF while maintaining view-consistency.}
\label{fig:method-overview}
\end{figure*}

\paragraph{Overview.} We present \ours, a geometry-consistent enhancer for NeRF models which substantially improves the visual quality and fidelity of rendered images. We show an overview of our method in Figure \ref{fig:method-overview}. Our proposed approach consists of three core elements: 3D Alignment, Non-Rigid Refinement and Geometric Attention. We leverage the fact the NeRF model has learned an estimate of the scene geometry and can render depth maps as well as RGB images. For a novel test view, we use depth maps and camera poses to 3D-align training image features to the novel camera viewpoint. To compensate for any slight inaccuracies in estimated geometry, we improve the alignment further with non-rigid refinement by means of a lightweight iterative optical flow network. We then regulate the contribution of any remaining misaligned regions with a geometry-aware attention module. Finally the image features are fused and processed with a Uformer \cite{wang22-uformer}, a 2D enhancer on which our method is based. We pre-train our model on a small dataset of render-GT image pairs and show that we can quickly fine-tune on a novel scene to achieve a substantial improvement in image quality.

\subsection{Preliminaries} \label{subsection:prelims}
NeRFs are trained on a set of ground truth RGB training images $\{H_{i}\}_{i=1}^{M}$ which capture a real world scene from set of known camera poses $\{C_{i}\}_{i=1}^{M}$, where $M$ is the size of the training set. After training, NeRFs are able to freely render RGB images $I_{i}$ and depths $D_{i}$ of the scene from any camera viewpoint, including novel camera poses $C_k$ not in $\{C_{i}\}_{i=1}^{M}$. NeRFs can interpolate well between camera poses, but the rendered images $I_{i}$ are typically degraded in quality compared to the ground truth $H_{i}$ especially in parts of the scene that contain high-frequency textures.

\subsection{Nearest Neighbour Selection} \label{subsection:nearestneighbours}

The high-frequency textures lost during the NeRF optimization process are still present in the set of training images used to train the NeRF. We take advantage of this by finding the set of training images which have the largest overlap of image content with the novel rendered image $I_{k}$. Given a novel camera pose $C_{k}$ and training poses $\{C_{i}\}_{1}^{M}$, we compute the nearest camera poses to $C_{k}$. We define a camera pose $C$ to be the concatenation of a rotation matrix $R$ and a translation vector $\bm{t}$ which describe the orientation and position of the camera respectively. To find the distance between rotations of different camera poses, we compute the three dimensional Euler angles of the rotation matrices and compute the mean of the L1 norm of the differences as follows:

\begin{equation}
dist^{i}_{ang}(R_{k}, R_{i}) = \frac{1}{3} \lvert \pi(R_{k})  - \pi(R_{i}) \rvert_{1} ,
\end{equation}
where $\pi(R_{i})$ are the Euler angles of the rotation matrix of $C_{i}$ and $dist^{i}_{ang}$ is the angular distance between $C_{k}$ and $C_{i}$, and $\lvert \; \rvert_{1}$ is the L1 norm. We also compute the distance between the camera positions as follows:

\begin{equation}
dist^{i}_{pos}(\bm{t}_{k}, \bm{t}_{i}) = \frac{1}{3} \lvert \bm{t}_{k}  - \bm{t}_{i} \rvert_{1} ,
\end{equation}
where $dist^{i}_{pos}$ is the positional distance between $C_{k}$ and $C_{i}$. Empirically and qualitatively, we find $dist^{i}_{pos}$ to be the strongest indicator for image content overlap with a novel view, followed second by $dist^{i}_{ang}$. We first choose the $5$ camera poses which have the smallest $dist^{i}_{pos}$ values, $\{C_{p}\}_{p=1}^{5}$, as follows:

\begin{equation}
\{C_{p}\}_{p=1}^{5} = \{C_{j}\} \mid \ j \in min_{5}\{dist^{j}_{pos}(\bm{t}_{k}, \{\bm{t}_j\}_{j=1}^M)\},
\end{equation}
where $min_{5}\{\} $ are the $5$ camera poses corresponding to the smallest distance values. Of these we choose the $n$ with the smallest $dist^{i}_{ang}$ values, $\{C_{i}\}_{i=1}^{n}$, as follows:

\begin{equation}
\{C_{i}\}_{i=1}^{n} = \{C_{p}\} \mid \ p \in min_{n}\{dist^{p}_{ang}(R_{k},\{R_p\}_{p=1}^5)\},
\end{equation}
where $min_{n}\{\}$ are the $n$ camera poses corresponding to the smallest distance values. The $n$ training images corresponding to these camera poses, $\{H_{i}\}_{i=1}^{n}$ are considered to be the closest neighbouring training images with respect to our novel camera view. 

\subsection{3D Alignment} \label{subsection:FR}

Once nearest neighbours are selected, we extract full resolution 64-dimensional image features using a small convolutional encoder block and reproject the features into the novel camera view using the pinhole camera model. This allows our enhancer to leverage relevant information from neighbours even if the cameras poses have very different orientation and position. Given a rendered novel view $I_{k}$ and a set of neighbouring training images $\{H_{i}\}_{i=1}^{n}$, we extract image features with two separate convolutional encoder blocks as follows:

\begin{equation}
I^{f}_{k} = conv_I(I_{k}),
\end{equation}

\begin{equation}
\{H^{f}_{i}\}_{i=1}^{n} = conv_H(\{H_{i}\}_{i=1}^{n}),
\end{equation}
where $I^{f}_{k}$ and $\{H^{f}_{i}\}_{i=1}^{n}$ are image features extracted from $I_{k}$ and $\{H_{i}\}_{i=1}^{n}$ respectively and $conv_I()$ and $conv_H()$ are small convolutional blocks. We reproject the neighbouring image features into the novel camera view $C_{k}$ using the pinhole camera model. Given a 3D coordinate $\bm{x}_{k} = (x_{k},y_{k},z_{k})$ in the novel camera view, where $(x_{k},y_{k})$ are the pixel coordinates in the image, and $z_{k}$ is the depth value at that pixel coordinate, we reproject the coordinate into a neighbouring camera view as follows:

\begin{equation}
\bm{x}_{k\rightarrow i} = K_{i}C_{i}C^{-1}_{k}K^{-1}_{k} [\bm{x}_{k},1],
\end{equation}
where $\bm{x}_{k\rightarrow i} = (x_{k\rightarrow i},y_{k\rightarrow i},z_{k\rightarrow i}, 1) $ is the 3D coordinate reprojected from camera $k$ to camera $i$, $K_{i}$ is the camera intrinsic matrix, $C^{-1}_{i}$ is the inverse of the camera pose and $[\;,\; ]$ denotes concatenation. To ensure geometric consistency, we conduct visibility testing by comparing the reprojected depth value $z_{k\rightarrow i}$ with the depth value computed by NeRF, $z_{i}$, as follows:

\begin{equation}
vis_{i} = \mathcal{H}\left(1 - \frac{z_{k\rightarrow i}}{z_{i}} + l\right)
\end{equation}
where $vis_{i}$ is the visibility score for the given pixel, $\mathcal{H}()$ denotes the Heaviside function, and $l$ is a leniency threshold to account for a degree of inaccuracy in depth and camera pose estimates. The reprojected feature map is formed by copying the values from the reprojected coordinates $\bm{x}_{k\rightarrow i}$ into the original coordinates $\bm{x}_{k}$, weighted by the visibility score:

\begin{equation}
H^{f}_{i\rightarrow k} \langle \phi(\bm{x}_{k}) \rangle = vis_{i} \times H^{f}_{i} \langle \phi(\bm{x}_{k\rightarrow i}) \rangle,
\end{equation}
where $H^{f}_{i\rightarrow k}$ are the image features reprojected from camera $i$ to camera $k$, $\phi(\bm{x}_{k}) = (\frac{x_{k}}{z_{k}}, \frac{y_{k}}{z_{k}})$ and $ \langle \; \rangle $ denotes 2D coordinate indexing.

\subsection{Non-Rigid Refinement} \label{subsection:IFA}

In real world data, there are often errors in camera pose estimation due to the limitations of COLMAP \cite{schoenberger16mvs, jiang23}, and also in the geometry estimated by NeRF, hence our 3D alignment is unlikely to find perfect correspondences. To account for this, we introduce a lightweight iterative optical flow network which further refines the alignment between the neighbouring images and the novel view image. Typically optical flow methods expect two clean images to accurately find correspondences but the domain gap between the blurry rendered image and the neighbouring images violates this assumption. Our choice however is motivated by the fact that optical flow methods can produce reasonable results based on global structures and shapes alone \cite{jiang23}. We use the flow network presented in \cite{catleychandar22} as it has been shown to work well even with domain gaps. We perform the iterative refinement in feature space and learn a network trained end-to-end which is optimized for our specific task instead of general purpose alignment \cite{kalantari19, catleychandar22}. Given the 3D aligned neighbouring features $H^{f}_{i\rightarrow k}$, we refine the alignment further as follows:

\begin{equation}
f_{H^{f}_{i\rightarrow k}} =  \mathcal{F}(H^{f}_{i\rightarrow k}, I^{f}_{k}),
\end{equation}

\begin{equation}
H^{f^{'}}_{i\rightarrow k} =  warp_{2D}(H^{f}_{i\rightarrow k}, f_{H^{f}_{i\rightarrow k}}),
\end{equation}
where $f_{H^{f}_{i\rightarrow k}}$ is the estimated flow field, $H^{f^{'}}_{i\rightarrow k}$ are the reprojected and warped neighbouring image features, $\mathcal{F}()$ is our lightweight optical flow network and $warp()_{2D}$ denotes the function for warping an image with a 2D flow field.

\subsection{Geometry-Aware Attention} \label{subsection:GA}
Any regions which remain misaligned after our combined 3D and 2D alignment can feed through to the final enhanced results as ghosting artifacts. Spatial attention has been shown to be effective at reducing such artifacts \cite{yan19}. We propose a learnable combined spatial and geometric attention module to regulate misaligned regions. Our geometry-aware attention is based on both 2D and 3D spatial information. This allows our enhancer to regulate contributions from neighbours based on similarities in pixel content and also geometric distance based on camera orientation and depths.  Given the neighbouring and novel view image features, as well as the neighbour depth projected to the novel view $D_{i\rightarrow k}$ and the novel view depth $D_{k}$, we compute the pixel attention weights as follows:

\begin{equation}
 \psi_{pix}^{i} = \mathcal{A}_{pix}(H^{f^{'}}_{i\rightarrow k}, I^{f}_{k}, D_{i\rightarrow k}, D_{k}),
\end{equation}
where $\psi_{pix}^{i} \in \mathbb{R}^{w\times h}$ are the pixel attention weights, $w$ and $h$ are the width and height of the  image and $\mathcal{A}_{pix}()$ is the pixel attention module which is composed of two convolutional layers and a sigmoid activation. In the second stage, the camera attention weights are computed using the Euler angles and positions of neighbour and novel view camera poses:

\begin{equation}
 \psi_{cam}^{i} = \mathcal{A}_{cam}(\pi(R_{i}), \pi(R_{k}), \mathbf{t}_{i}, \mathbf{t}_{k})
\end{equation}
where $ \psi_{cam}^{i} \in \mathbb{R}^{1}$ are the camera attention weights and $\mathcal{A}_{cam}()$ is the camera attention module which is composed of two fully connected layers and a sigmoid activation. Finally both sets of the weights are applied to the neighbour image features. $\psi_{pix}$ is applied at a per-pixel level while $\psi_{cam}$ is applied at a per-image level:

\begin{equation}
H^{f^{a}}_{i\rightarrow k} = \psi_{cam}^{i} \times \psi_{pix}^{i} \times H^{f^{'}}_{i\rightarrow k},
\end{equation}
where $H^{f^{a}}_{i\rightarrow k}$ are the attention regulated image features.

\subsection{Feature Fusion} \label{subsection:FF}

We use the maxpooling approach described in \cite{aittala18} to combine our set of attention regulated neighbouring features, $\{H^{f^{a}}_{i\rightarrow k}\}_{i=1}^n$ into a single feature map $H^{f}_{pool}$. This approach has the advantages of outperforming concatenation \cite{catleychandar22} and also defining a flexible architecture which can accept any number of input neighbours. This gives us the ability to use fewer or more neighbouring images depending on factors such as availability of data, image resolution and GPU memory, ensuring our enhancer is practical and can be applied in multiple settings. Finally we process the pooled features and novel view features together with a convolutional layer and enhance them further using a 2D enhancer, Uformer \cite{wang22-uformer}, which we modify to accept image features instead of RGB inputs:

\begin{equation}
\hat{H}_{k} = \mathcal{U}(conv_{merge}([I^{f}_{k}, H^{f}_{pool}])),
\end{equation}
where $\mathcal{U}$ is the Uformer, and $conv_{merge}()$ denotes a convolutional layer.

\subsection{Pre-training and Implementation Details} \label{subsection:training}

Our enhancer is first pre-trained using a single NeRF baseline and dataset, specifically NeRF \cite{mildenhall20} and LLFF \cite{mildenhall19}. We then fine-tune for 1 hour on each new scene or novel NeRF baseline method. To generate the training and fine-tuning data for our enhancer, we first train a given baseline NeRF model on a scene and render all images from the training set, which generates a set of render-GT pairs. We pre-train our model on all scenes from the LLFF dataset using the renders generated by NeRF \cite{mildenhall20} for 3000 epochs (approximately 5 days) and fine-tune on novel scenes and NeRF models for one hour per scene, which is comparable to the per-scene training time of state-of-the-art NeRF methods \cite{barron23}.  We reduce L1 and perceptual losses between the enhanced image and ground truth as follows:

\begin{equation}
L = |\hat{H}_{i} - H_i|_{1} + 10^{-3}|\omega(\hat{H}_{i}) - \omega(H_i)|_{1},
\end{equation}
where $\hat{H}_{i}$ is our predicted enhanced image, $H_i$ is the GT, $L$ is our loss function and $\omega()$ is a pre-trained VGG-19 \cite{simonyan15}. We use random crops of size 512$\times$512 with a batch size of 4 and a learning rate of $1\times10^{-4}$ with the Adam optimizer \cite{kingma14}. We use a leniency threshold of 0.25 for visibility testing and we use 5 neighbours for the LLFF and 360v2 datasets, and 2 neighbours for the DTU dataset. We train our model using PyTorch \cite{pytorch, torchvision} on 4$\times$ NVidia V100 GPUs. For all baselines and enhancers, we use the official code and checkpoints provided by authors when available. For reproducibility, we provide full implementation details of each component of our method in the supplementary.

\section{Results}

\subsection{Datasets and Metrics}

 We evaluate our method on three varied real world multi-view datasets, LLFF (8 scenes) \cite{mildenhall19}, 360v2 (9 scenes) \cite{barron22} and DTU (124 scenes) \cite{jensen14}. Together, these datasets contain a mixture of front-facing and 360$\degree$ scenes, both indoor bounded and outdoor unbounded, with complex geometries and a range of high-frequency textures. The total number of images per scene varies from 20 to 311. Note, we evaluate all scenes from LLFF and 360v2 at a consistent 4$\times$ downsampled resolution, unlike previous works which evaluate the indoor and outdoor scenes from 360v2 at different downsampling rates. We evaluate all DTU scenes at full resolution and use the diffuse light setting. For each scene, every eighth image is held out for testing and the remaining images are used to train the NeRF baselines and fine-tune our enhancer. Following previous works, we use PSNR, SSIM \cite{wang04} and LPIPS (VGG) \cite{zhang18} to evaluate our method. 

\subsection{Baseline Methods}

To demonstrate the general applicability of our proposed enhancer, we extensively evaluate our method using six different baseline NeRF methods which together represent the evolution of neural rendering over the last few years. These include the current state-of-the-art with respect to image fidelity (MipNeRF360 \cite{barron22}, ZipNeRF \cite{barron23}), training speed (Nerfacto \cite{tancik23}, NeuS2++ an unbounded variant of NeuS2 \cite{wang22-neus2}) and older seminal works (NeRF \cite{mildenhall20}, TensoRF \cite{chen22}). We compare to state-of-the-art NeRF enhancers including Boosting View Synthesis \cite{rong22}, NeRFLiX \cite{zhou23} and NeRFLiX++ \cite{zhou23pami}, an extension of NeRFLiX using a GAN-based degradation simulator. For each dataset, we report results averaged across all scenes in Table \ref{table:all-results}. We provide per-scene results of each method and also compare to AligNeRF \cite{jiang23} on outdoor scenes in the supplementary. As there is no available code, we report results directly from the original works for \cite{jiang23}, \cite{rong22} and \cite{zhou23pami}.

\begin{table}[t]
\centering
\caption{Quantitative evaluation of our enhancer applied to six different NeRF baselines. Results are averaged across all scenes for each dataset. Red and orange highlights indicate 1st and 2nd best performing methods. Our model consistently outperforms all baselines and other enhancers across all metrics. $^\dagger$Results as reported by authors.}
\begin{adjustbox}{width=0.99\textwidth}
\begin{tabular}{l c c c c} \\ \toprule
  Model & Dataset & PSNR (dB) $\uparrow$  & SSIM$\uparrow$ & LPIPS$\downarrow$ \\ \midrule  

    ZipNeRF  & 360v2 & \hspace{-1.2cm}28.90  &\hspace{-0.87cm} \cellsecond0.8367 &\hspace{-0.97cm} \cellsecond 0.1779 \\ 
    ZipNeRF + NeRFLiX &  360v2 & \cellsecond 29.00  ($\uparrow$ 0.10) & \hspace{0.5cm} 0.8317 ($\downarrow$ 0.005) & \hspace{0.5cm} 0.2045 ($\uparrow$ 15.0\%) \\
    ZipNeRF + Ours   &  360v2 & \cellbest 29.23 ($\uparrow$ 0.33) & \hspace{0.5cm} \cellbest 0.8465 ($\uparrow$ 0.098)  & \cellbest\hspace{0.35cm}  0.1662 ($\downarrow$ 6.6\%) \\ 
     \midrule 
     
    MipNeRF360  &  360v2 &\hspace{-1.2cm}28.26 &\hspace{-0.87cm} \cellsecond 0.8050 &\hspace{-0.97cm} \cellsecond 0.2297 \\
    MipNeRF360 + NeRFLiX &  360v2 & \cellsecond 28.44 ($\uparrow$ 0.18) &\hspace{0.5cm} 0.8036 ($\downarrow$ 0.001)&\hspace{0.35cm} 0.2441 ($\uparrow$ 6.3\%)  \\
    MipNeRF360 + Ours   &  360v2 & \cellbest 28.89 ($\uparrow$ 0.63) &\hspace{0.5cm} \cellbest 0.8302 ($\uparrow$ 0.025) &\hspace{0.5cm} \cellbest 0.1987 ($\downarrow$ 13.5\%)\\  \midrule  
    
    Nerfacto  &  360v2 & \hspace{-1.2cm}26.11& \hspace{-0.87cm} 0.7157 &\hspace{-0.97cm} 0.3266 \\
    Nerfacto + NeRFLiX &  360v2 & \cellsecond 26.92($\uparrow$ 0.81)&\hspace{0.5cm} \cellsecond 0.7410 ($\uparrow$ 0.025)&\hspace{0.35cm} \cellsecond 0.3044 ($\downarrow$ 6.8\%)  \\
    Nerfacto + Ours & 360v2 & \cellbest 27.45($\uparrow$ 1.34)&\hspace{0.5cm} \cellbest 0.7700 ($\uparrow$ 0.054)&\hspace{0.5cm} \cellbest 0.2670  ($\downarrow$ 18.2\%) \\  \midrule  
    
    NeuS2++ &  DTU &\hspace{-1.2cm}27.35 & \hspace{-0.87cm} 0.7587&\hspace{-0.97cm} 0.4386 \\
    NeuS2++ + NeRFLiX &  DTU &\cellsecond 27.40 ($\uparrow$ 0.05)&\hspace{0.5cm}  \cellsecond0.7752 ($\uparrow$ 0.017)& \hspace{0.35cm} \cellsecond0.4167 ($\downarrow$ 5.0\%)\\
    NeuS2++ + Ours   &  DTU &\cellbest 28.46 ($\uparrow$ 1.11) &\hspace{0.5cm}  \cellbest 0.8237 ($\uparrow$ 0.065)&\hspace{0.5cm} \cellbest0.3939 ($\downarrow$ 10.2\%)\\  \midrule   
    
    NeRF & LLFF & \hspace{-1.2cm}26.57 & \hspace{-0.87cm} 0.8170 & \hspace{-0.97cm} 0.2389 \\
    NeRF + Boosting View Synthesis$^\dagger$  & LLFF &27.08 ($\uparrow$ 0.51) &\hspace{0.5cm} 0.8371 ($\uparrow$ 0.020)& - \\ 
    NeRF + NeRFLiX  & LLFF &27.17 ($\uparrow$ 0.60)&\hspace{0.5cm}  0.8552 ($\uparrow$ 0.038)&\hspace{0.5cm}  \cellsecond0.1695 ($\downarrow$ 29.0\%)\\ 
    NeRF + NeRFLiX++$^\dagger$  & LLFF &\cellsecond27.25 ($\uparrow$ 0.68)&\hspace{0.5cm}  \cellsecond0.8580 ($\uparrow$ 0.041)& - \\ 
    NeRF + Ours  & LLFF &\cellbest 27.67 ($\uparrow$ 1.10)  &\hspace{0.5cm}  \cellbest0.8713 ($\uparrow$ 0.054)&\hspace{0.5cm}  \cellbest0.1495 ($\downarrow$ 37.4\%)\\  \midrule   
    
    TensoRF & LLFF &\hspace{-1.2cm}26.88 & \hspace{-0.87cm} 0.8432 & \hspace{-0.97cm} 0.1829 \\
    TensoRF + NeRFLiX  & LLFF &\cellsecond27.38 ($\uparrow$ 0.50)&\hspace{0.5cm} 0.8652 ($\uparrow$ 0.022)& \hspace{0.5cm} \cellsecond0.1514 ($\downarrow$ 17.2\%)\\ 
    TensoRF + NeRFLiX++$^\dagger$  & LLFF &\cellsecond27.38 ($\uparrow$ 0.50)& \hspace{0.5cm} \cellsecond0.8660 ($\uparrow$ 0.023) & - \\ 
    TensoRF + Ours  & LLFF &\cellbest27.58 ($\uparrow$ 0.70)& \hspace{0.5cm} \cellbest0.8670 ($\uparrow$ 0.024) & \hspace{0.5cm} \cellbest0.1494 ($\downarrow$ 18.3\%)\\ \bottomrule 

\end{tabular}
\end{adjustbox}
\label{table:all-results}
 \end{table}



\subsection{Quantitative and Qualitative Evaluation}

\begin{figure}[hp]
\centering
\includegraphics[width=1.0\textwidth]{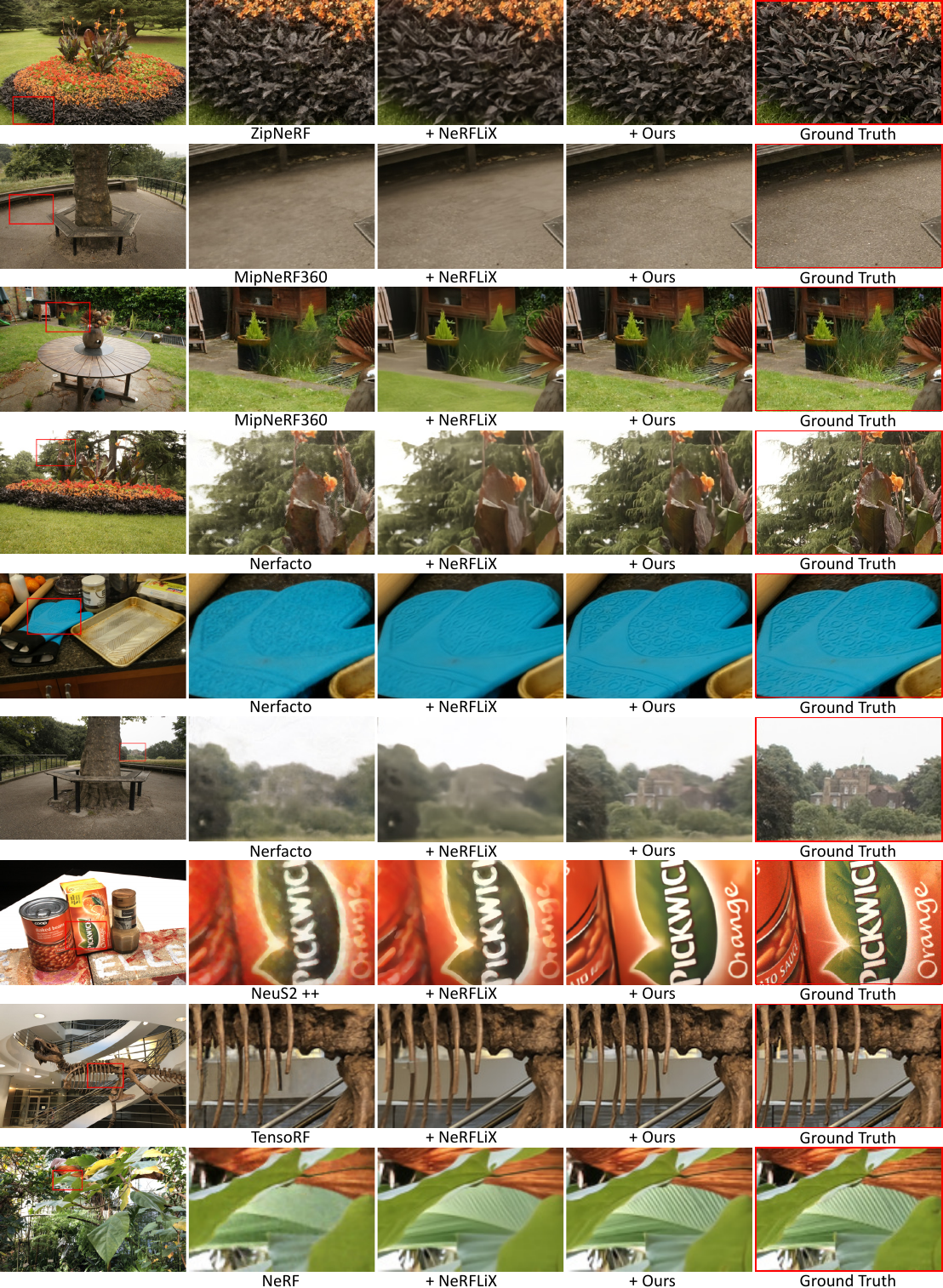}
\caption{Qualitative comparisons with six different NeRF baseline models across three datasets. Our method recovers more detail in high-frequency regions such as foliage, tarmac floor, patterns on the glove, the edges of the building and text.}
\label{fig:results-fig-eccv}
\end{figure}

We present quantitative results in Table \ref{table:all-results}. We show that our model achieves improvements in all metrics for all six NeRF baselines.  On the 360v2 dataset, \ours~improves the PSNR of Nerfacto by 1.34dB, MipNeRF360 by 0.63dB and ZipNeRF by 0.33dB, and achieves corresponding reductions in LPIPS of 18.2\%, 13.5\% and 6.6\% respectively. On the DTU and LLFF datasets, \ours~improves the PSNR of Neus2++ by 1.11dB, NeRF by 1.10dB and TensoRF by 0.70dB, with corresponding reductions in LPIPS of 10.2\%, 37.4\% and 18.3\% respectively. We consistently improve SSIM scores for all baselines and datasets. We also outperform state-of-the-art NeRF enhancers NeRFLiX and NeRFLiX++ in every setting. These results are reflected in the qualitative evaluation shown in Figures \ref{fig:teaser-fig} and \ref{fig:results-fig-eccv}. We note that ZipNeRF already achieves very high fidelity in high-frequency regions, so the improvements from our enhancer are largely from denoising without over-smoothing. We show noticeable improvements over all other NeRF baseline models and NeRFLiX in restoring high-frequency details. We also present video results in the supplementary which demonstrate that \ours \ has view-consistency in line with the baseline NeRF models, as opposed to NeRFLiX which is geometry-agnostic and suffers from flickering in high-frequency regions.

 \begin{table}[t]
\centering
\caption{Evaluation of robustness to camera pose noise. We present MUSIQ \cite{ke21} scores when Gaussian noise is added to training camera poses on the Garden scene from the 360v2 dataset. Small, medium and large noise settings correspond to standard deviations of 6.25e-2/3.125e-4 | 12.5e-2/6.25e-4 |  25e-2/12.5e-4 with respect to camera rotation/position. }
\begin{adjustbox}{width=0.85\textwidth}
\begin{tabular}{l  c  c  c  c } \\ \toprule
  Model  & \hspace{0.5cm} No Noise  & | Small Noise & | Medium Noise  & | Large Noise \\ \midrule
   
   Nerfacto   & \cellsecond\hspace{-0.92cm}66.50  & \cellsecond\hspace{-0.92cm}53.33 & \cellsecond\hspace{-1.1cm}41.65 & \cellsecond\hspace{-1.1cm}32.43  \\
    Nerfacto + Ours & \cellbest72.32 ($\uparrow$5.8)  & \cellbest62.59 ($\uparrow$9.3) & \cellbest58.94 ($\uparrow$17.3) & \cellbest51.60 ($\uparrow$19.2)\\ \midrule
    ZipNeRF   & \cellsecond\hspace{-0.92cm}71.38  & \cellsecond\hspace{-0.92cm}63.29&\cellsecond \hspace{-0.92cm}58.67 &\cellsecond \hspace{-0.92cm}49.50  \\
    ZipNeRF + Ours & \cellbest72.06 ($\uparrow$0.7)  & \cellbest69.76 ($\uparrow$6.5)& \cellbest65.03  ($\uparrow$6.4)& \cellbest52.39($\uparrow$2.9)\\ \bottomrule

\end{tabular}
\end{adjustbox} 
\label{table:ablation-noise}
 \end{table}

\subsection{Robustness To Inaccurate Camera Calibration}

\begin{figure}[h]
\centering
\includegraphics[width=0.82\textwidth]{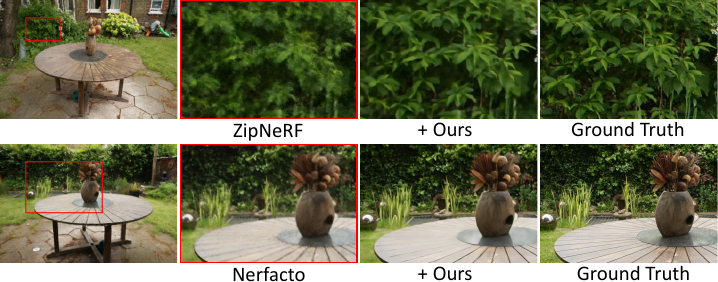}
\caption{Qualitative results when adding medium pose noise. The quality of the baseline suffers greatly while our enhancer is robust and achieves noticeable improvements.}
\label{fig:results-noise}
\end{figure}

We evaluate the robustness of our method to inaccurate camera calibration in Table \ref{table:ablation-noise}. To simulate the effects of inaccurate camera pose estimation, we apply increasing levels of additive zero mean Gaussian noise to the camera poses estimated by COLMAP. The rotation standard deviation is expressed in degrees while position standard deviation is expressed as a unit of physical distance, with total scene size set to a bounding cube of length 2. For each noise level, we train Nerfacto and ZipNeRF using the noisy camera poses and evaluate the ability of our enhancer to improve the perceptual quality of the noisy NeRF baselines. Training with incorrect camera poses introduces pixel shifts between rendered images and ground truth, so we assess performance with MUSIQ \cite{ke21}, a SOTA no-reference image quality assessment metric. We show that our improvements over Nerfacto and ZipNeRF are larger when using noisy camera poses compared to the no noise setting. This is reflected in Figure \ref{fig:results-noise} where we qualitatively evaluate the medium noise setting. The baselines suffer a large drop in image quality while \ours~is robust to inaccurate poses and is able to achieve noticeable improvements over the noisy baselines.


\subsection{Ablation Study}

\begin{figure}[!h]
\centering
\includegraphics[width=0.82\textwidth]{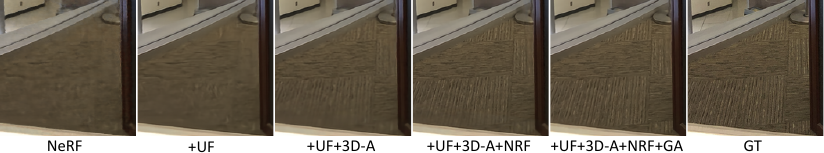}
\caption{Qualitative results of the ablation study. Settings correspond to Table \ref{table:ablation-all}.}
\label{fig:ablation_rebuttal}
\end{figure}

We validate the importance of individual components of our method in Table \ref{table:ablation-all}. We conduct the study across all scenes in the LLFF dataset and present average results. Our combined contributions show an improvement of 1.1dB PSNR over NeRF and 0.93dB over a strong 2D baseline enhancer, Uformer \cite{wang22-uformer} on which our proposed approach is based. We show a sharp improvement in LPIPS perceptual quality from our 3D alignment module (7.4\%), and a further improvement from our non-rigid refinement (13.4\%), which greatly improves the ability of our model to find accurate correspondences and correct errors in geometry. We show qualitative results of our ablation in Figure \ref{fig:ablation_rebuttal}. We evaluate the following model components: \textbf{UF}: Baseline 2D enhancer Uformer which our approach is based on.  \textbf{NN}: Nearest Neighbour Selection. \textbf{3D-A}: Our 3D Alignment module comprising of feature reprojection using depths and camera poses. \textbf{NRF}: Our Non-Rigid Refinement module comprising of an iterative lightweight optical flow network which further aligns the reprojected features. \textbf{GA}: Our Geometric Attention module comprising a geometry-aware feature regulation mechanism. \textbf{PT}: The effect of pre-training our model. We conduct a more detailed ablation study on pre-training and fine-tuning in Table \ref{table:ablation-finetune}. Here we show two things; that our method can achieve large improvements in as little as 1 minute of fine-tuning; and that it is our complete combined contributions which allow our method to learn novel image degradations so quickly. Uformer+NN also has access to nearest neighbours from the training set, but as it is unaware of geometry, it is not capable of quickly leveraging the information from nearby viewpoints. 

  \begin{table}[t]
\centering
\caption{Ablation on the performance of our individual contributions averaged across all eight scenes from the LLFF dataset with NeRF as the baseline model.}
\begin{adjustbox}{width=0.80\textwidth}
\begin{tabular}{l c c c c c c | c c c} \\ \toprule
     Model & UF & NN & 3D-A & NRF & GA & PT & PSNR & SSIM & LPIPS  \\ \midrule
      NeRF & $\square$ &   $\square$ & $\square$  & $\square$ & $\square$ & $\square$ & 26.57& 0.8170& 0.2389 \\
    + Uformer  &\hspace{0.045cm}\boxtick  & $\square$ & $\square$ & $\square$ & $\square$ &  $\square$ & 26.74& 0.8314& 0.2245 \\
     + 3D Alignment&\hspace{0.045cm}\boxtick  &  \hspace{0.045cm}\boxtick & \hspace{0.045cm}\boxtick & $\square$& $\square$ &$\square$  & 26.84  & 0.8397 & 0.2078 \\
     + Non-Rigid Refinement  &\hspace{0.045cm}\boxtick  &  \hspace{0.045cm}\boxtick & \hspace{0.045cm}\boxtick & \hspace{0.045cm}\boxtick & $\square$ &$\square$& 26.93 & 0.8467 & 0.1799  \\
  + Geometric Attention&\hspace{0.045cm}\boxtick &  \hspace{0.045cm}\boxtick & \hspace{0.045cm}\boxtick & \hspace{0.045cm}\boxtick & \hspace{0.045cm}\boxtick   & $\square$ & \cellsecond 27.00  &\cellsecond 0.8501 &\cellsecond 0.1755 \\
    + Pre-training  &\hspace{0.045cm}\boxtick &  \hspace{0.045cm}\boxtick & \hspace{0.045cm}\boxtick & \hspace{0.045cm}\boxtick & \hspace{0.045cm}\boxtick & \hspace{0.045cm}\boxtick  &\cellbest 27.67  & \cellbest0.8713 & \cellbest0.1495\\
\midrule
\end{tabular}
\end{adjustbox} 
\label{table:ablation-all}
 \end{table}

  \begin{table}[t]
\centering
\caption{Ablation on fine-tuning time on the Garden scene from the 360v2 dataset with Nerfacto as the baseline model. Column headers on the right indicate time spent fine-tuning each model. Results presented are PSNR (dB).}
\begin{adjustbox}{width=0.75\textwidth}
\begin{tabular}{l c c c c c c | c c c} \\ \toprule
   Model & UF & NN & 3D-A & NRF & GA & PT & 1 Min & 10 Minutes & 1 Hour \\ \midrule
    Nerfacto &$\square$ & $\square$ &$\square$ & $\square$  & $\square$  & $\square$ & 25.28 & 25.28 & 25.28\\
    + Uformer & \hspace{0.045cm}\boxtick &$\square$ & $\square$  & $\square$  & $\square$  &\hspace{0.045cm}\boxtick & \cellsecond 25.43 & 25.52 &\cellsecond 25.73\\
    + Uformer+NN  & \hspace{0.045cm}\boxtick &\hspace{0.045cm}\boxtick  & $\square$  & $\square$  & $\square$ &\hspace{0.045cm}\boxtick & 25.39 &\cellsecond 25.60 & 25.67 \\
 + Ours & \hspace{0.045cm}\boxtick & \hspace{0.045cm}\boxtick  & \hspace{0.045cm}\boxtick   & \hspace{0.045cm}\boxtick  & \hspace{0.045cm}\boxtick  &\hspace{0.045cm}\boxtick &\cellbest  26.08 &\cellbest  26.19 & \cellbest 26.24\\
\midrule
\end{tabular}
\end{adjustbox} 
\label{table:ablation-finetune}
 \end{table}

\noindent \textbf{Limitations} A limitation of our method is the requirement to store all relevant training images. This could be prohibitive for very large scenes, especially at higher image resolutions. Secondly, although we are several times faster than the baseline NeRF methods, our approach does not yet achieve real time (\ie 30fps) inference. \textbf{Societal Impact} As NeRFs become more editable, our method could be misused to improve the photorealism of generated videos of people, places and objects. Currently our method requires ground truth images with a dense 3D coverage of the scene, so it is not trivial to use in the wild. 
\section{Conclusion}

We have presented \ours, a geometry-consistent NeRF enhancer which combines concepts from 3D and 2D vision to substantially improve the image quality of NeRF renderings in real world settings. Our model accurately finds correspondences between different camera views by performing 3D alignment and non-rigid refinement, while also being robust to errors in camera pose estimation and reducing reprojection artifacts with geometry-aware attention. \ours \ achieves consistent improvements in image quality over six varied NeRF baselines and existing NeRF enhancers across three challenging real world datasets. Our model demonstrates wide applicability and strong generalization, fine-tuning on a novel scene in under $60$ minutes to learn the distribution of image degradations.

%
%

\bibliographystyle{splncs04}
\bibliography{main}

\begin{thebibliography}{10}
\providecommand{\url}[1]{\texttt{#1}}
\providecommand{\urlprefix}{URL }
\providecommand{\doi}[1]{https://doi.org/#1}

\bibitem{aittala18}
Aittala, M., Durand, F.: Burst image deblurring using permutation invariant convolutional neural networks. In: ECCV (September 2018)

\bibitem{barron21}
Barron, J.T., Mildenhall, B., Tancik, M., Hedman, P., Martin-Brualla, R., Srinivasan, P.P.: Mip-nerf: A multiscale representation for anti-aliasing neural radiance fields. In: ICCV. pp. 5835--5844 (2021)

\bibitem{barron22}
Barron, J.T., Mildenhall, B., Verbin, D., Srinivasan, P.P., Hedman, P.: Mip-nerf 360: Unbounded anti-aliased neural radiance fields. In: CVPR (2022)

\bibitem{barron23}
Barron, J.T., Mildenhall, B., Verbin, D., Srinivasan, P.P., Hedman, P.: Zip-nerf: Anti-aliased grid-based neural radiance fields. In: ICCV (2023)

\bibitem{bian23}
Bian, W., Wang, Z., Li, K., Bian, J.W., Prisacariu, V.A.: Nope-nerf: Optimising neural radiance field with no pose prior. In: CVPR. pp. 4160--4169 (June 2023)

\bibitem{catleychandar22}
Catley-Chandar, S., Tanay, T., Vandroux, L., Leonardis, A., Slabaugh, G., P\'erez-Pellitero, E.: Flex{HDR}: Modeling alignment and exposure uncertainties for flexible {HDR} imaging. IEEE TIP  \textbf{31} (2022)

\bibitem{chen22}
Chen, A., Xu, Z., Geiger, A., Yu, J., Su, H.: Tensorf: Tensorial radiance fields. In: ECCV (2022)

\bibitem{dhamo23}
Dhamo, H., Nie, Y., Moreau, A., Song, J., Shaw, R., Zhou, Y., P{\'e}rez-Pellitero, E.: Headgas: Real-time animatable head avatars via 3d gaussian splatting. arXiv preprint arXiv:2312.02902  (2023)

\bibitem{huang23}
Huang, X., Li, W., Hu, J., Chen, H., Wang, Y.: Refsr-nerf: Towards high fidelity and super resolution view synthesis. In: CVPR. pp. 8244--8253. IEEE Computer Society, Los Alamitos, CA, USA (jun 2023)

\bibitem{isik23}
I\c{s}{\i}k, M., Rünz, M., Georgopoulos, M., Khakhulin, T., Starck, J., Agapito, L., Nießner, M.: Humanrf: High-fidelity neural radiance fields for humans in motion. ACM TOG  \textbf{42}(4),  1--12 (2023)

\bibitem{jensen14}
Jensen, R., Dahl, A., Vogiatzis, G., Tola, E., Aan{\ae}s, H.: Large scale multi-view stereopsis evaluation. In: CVPR. pp. 406--413. IEEE (2014)

\bibitem{jiang23}
Jiang, Y., Hedman, P., Mildenhall, B., Xu, D., Barron, J.T., Wang, Z., Xue, T.: Alignerf: High-fidelity neural radiance fields via alignment-aware training. In: CVPR. pp. 46--55 (2023)

\bibitem{kalantari19}
Kalantari, N.K., Ramamoorthi, R.: {Deep HDR Video from Sequences with Alternating Exposures}. Comput. Graph. Forum  (2019)

\bibitem{ke21}
Ke, J., Wang, Q., Wang, Y., Milanfar, P., Yang, F.: Musiq: Multi-scale image quality transformer. In: ICCV. pp. 5148--5157 (October 2021)

\bibitem{kingma14}
Kingma, D., Ba, J.: Adam: A method for stochastic optimization. In: ICLR (2015)

\bibitem{li22}
Li, T., Slavcheva, M., Zollhoefer, M., Green, S., Lassner, C., Kim, C., Schmidt, T., Lovegrove, S., Goesele, M., Newcombe, R.A., Lv, Z.: Neural 3d video synthesis from multi-view video. In: CVPR. pp. 5511--5521 (2021)

\bibitem{lin21}
Lin, C.H., Ma, W.C., Torralba, A., Lucey, S.: Barf: Bundle-adjusting neural radiance fields. In: ICCV (2021)

\bibitem{liu23}
Liu, Y.L., Gao, C., Meuleman, A., Tseng, H.Y., Saraf, A., Kim, C., Chuang, Y.Y., Kopf, J., Huang, J.B.: Robust dynamic radiance fields. In: CVPR (2023)

\bibitem{torchvision}
maintainers, T., contributors: {TorchVision: PyTorch's Computer Vision library} (Nov 2016)

\bibitem{meuleman23}
Meuleman, A., Liu, Y.L., Gao, C., Huang, J.B., Kim, C., Kim, M.H., Kopf, J.: Progressively optimized local radiance fields for robust view synthesis. In: CVPR. pp. 16539--16548 (June 2023)

\bibitem{mildenhall22}
Mildenhall, B., Hedman, P., Martin-Brualla, R., Srinivasan, P.P., Barron, J.T.: {NeRF} in the dark: High dynamic range view synthesis from noisy raw images. In: CVPR (2022)

\bibitem{mildenhall19}
Mildenhall, B., Srinivasan, P.P., Ortiz-Cayon, R., Kalantari, N.K., Ramamoorthi, R., Ng, R., Kar, A.: Local light field fusion: Practical view synthesis with prescriptive sampling guidelines. ACM TOG  (2019)

\bibitem{mildenhall20}
Mildenhall, B., Srinivasan, P.P., Tancik, M., Barron, J.T., Ramamoorthi, R., Ng, R.: Nerf: Representing scenes as neural radiance fields for view synthesis. In: ECCV (2020)

\bibitem{moreau21}
Moreau, A., Piasco, N., Tsishkou, D.V., Stanciulescu, B., de~La~Fortelle, A.: Lens: Localization enhanced by nerf synthesis. In: Conference on Robot Learning (2021)

\bibitem{moreau24}
Moreau, A., Song, J., Dhamo, H., Shaw, R., Zhou, Y., P{\'e}rez-Pellitero, E.: Human gaussian splatting: Real-time rendering of animatable avatars. In: CVPR (2024)

\bibitem{mueller22}
M\"uller, T., Evans, A., Schied, C., Keller, A.: Instant neural graphics primitives with a multiresolution hash encoding. ACM TOG  \textbf{41}(4) (2022)

\bibitem{niemeyer20}
Niemeyer, M., Geiger, A.: Giraffe: Representing scenes as compositional generative neural feature fields. In: CVPR (2021)

\bibitem{park23}
Park, K., Henzler, P., Mildenhall, B., Barron, J.T., Martin-Brualla, R.: Camp: Camera preconditioning for neural radiance fields. ACM Trans. Graph.  (2023)

\bibitem{pytorch}
Paszke, A., Gross, S., Massa, F., Lerer, A., Bradbury, J., Chanan, G., Killeen, T., Lin, Z., Gimelshein, N., Antiga, L., Desmaison, A., Kopf, A., Yang, E., DeVito, Z., Raison, M., Tejani, A., Chilamkurthy, S., Steiner, B., Fang, L., Bai, J., Chintala, S.: {PyTorch: An Imperative Style, High-Performance Deep Learning Library}. In: Wallach, H., Larochelle, H., Beygelzimer, A., d'Alché Buc, F., Fox, E., Garnett, R. (eds.) NeurIPS. pp. 8024--8035. Curran Associates, Inc. (2019)

\bibitem{peng21}
Peng, S., Dong, J., Wang, Q., Zhang, S., Shuai, Q., Zhou, X., Bao, H.: Animatable neural radiance fields for modeling dynamic human bodies. In: ICCV (2021)

\bibitem{peng23}
Peng, S., Yan, Y., Shuai, Q., Bao, H., Zhou, X.: Representing volumetric videos as dynamic mlp maps. In: CVPR (2023)

\bibitem{pumarola20}
Pumarola, A., Corona, E., Pons-Moll, G., Moreno-Noguer, F.: {D-NeRF: Neural Radiance Fields for Dynamic Scenes}. In: CVPR (2020)

\bibitem{raoult17}
Raoult, V., Reid-Anderson, S., Ferri, A., Williamson, J.E.: How reliable is structure from motion (sfm) over time and between observers? a case study using coral reef bommies. Remote Sensing  \textbf{9}(7) (2017)

\bibitem{roessle23}
Roessle, B., M{\"u}ller, N., Porzi, L., Bul{\`o}, S.R., Kontschieder, P., Nie{\ss}ner, M.: Ganerf: Leveraging discriminators to optimize neural radiance fields. ACM TOG  (2023)

\bibitem{rong22}
Rong, X., Huang, J.B., Saraf, A., Kim, C., Kopf, J.: Boosting view synthesis with residual transfer. In: CVPR (2022)

\bibitem{yu21}
{Sara Fridovich-Keil and Alex Yu}, Tancik, M., Chen, Q., Recht, B., Kanazawa, A.: Plenoxels: Radiance fields without neural networks. In: CVPR (2022)

\bibitem{schoenberger16sfm}
Sch\"{o}nberger, J.L., Frahm, J.M.: Structure-from-motion revisited. In: CVPR (2016)

\bibitem{schoenberger16mvs}
Sch\"{o}nberger, J.L., Zheng, E., Pollefeys, M., Frahm, J.M.: Pixelwise view selection for unstructured multi-view stereo. In: ECCV (2016)

\bibitem{shaw23}
Shaw, R., Song, J., Moreau, A., Nazarczuk, M., Catley-Chandar, S., Dhamo, H., Perez-Pellitero, E.: Swags: Sampling windows adaptively for dynamic 3d gaussian splatting. arXiv preprint arXiv:2312.13308  (2023)

\bibitem{simonyan15}
Simonyan, K., Zisserman, A.: Very deep convolutional networks for large-scale image recognition. In: ICLR (2015)

\bibitem{tanay23}
Tanay, T., Leonardis, A., Maggioni, M.: Efficient view synthesis and 3d-based multi-frame denoising with multiplane feature representations. In: CVPR (2023)

\bibitem{tanay24}
Tanay, T., Maggioni, M.: Global latent neural rendering. In: CVPR (2024)

\bibitem{tancik20ffn}
Tancik, M., Srinivasan, P.P., Mildenhall, B., Fridovich-Keil, S., Raghavan, N., Singhal, U., Ramamoorthi, R., Barron, J.T., Ng, R.: Fourier features let networks learn high frequency functions in low dimensional domains. In: NeurIPS (2020)

\bibitem{tancik23}
Tancik, M., Weber, E., Ng, E., Li, R., Yi, B., Kerr, J., Wang, T., Kristoffersen, A., Austin, J., Salahi, K., Ahuja, A., McAllister, D., Kanazawa, A.: Nerfstudio: A modular framework for neural radiance field development. In: ACM SIGGRAPH 2023 Conference Proceedings. SIGGRAPH '23 (2023)

\bibitem{truong23}
Truong, P., Rakotosaona, M.J., Manhardt, F., Tombari, F.: Sparf: Neural radiance fields from sparse and noisy poses. In: CVPR. pp. 4190--4200 (June 2023)

\bibitem{turki24}
Turki, H., Agrawal, V., Bulò, S.R., Porzi, L., Kontschieder, P., Ramanan, D., Zollh\"{o}fer, M., Richardt, C.: Hybridnerf: Efficient neural rendering via adaptive volumetric surfaces. In: Computer Vision and Pattern Recognition (CVPR) (2024)

\bibitem{verbin22}
Verbin, D., Hedman, P., Mildenhall, B., Zickler, T., Barron, J.T., Srinivasan, P.P.: {Ref-NeRF}: Structured view-dependent appearance for neural radiance fields. In: CVPR (2022)

\bibitem{wang22-clipnerf}
Wang, C., Chai, M., He, M., Chen, D., Liao, J.: Clip-nerf: Text-and-image driven manipulation of neural radiance fields. In: CVPR. pp. 3835--3844 (2022)

\bibitem{wang22-nerfsr}
Wang, C., Wu, X., Guo, Y.C., Zhang, S.H., Tai, Y.W., Hu, S.M.: Nerf-sr: High quality neural radiance fields using supersampling. In: ACM MM. p. 6445–6454. MM '22, Association for Computing Machinery, New York, NY, USA (2022)

\bibitem{wang22-neus2}
Wang, Y., Han, Q., Habermann, M., Daniilidis, K., Theobalt, C., Liu, L.: Neus2: Fast learning of neural implicit surfaces for multi-view reconstruction. In: ICCV (2023)

\bibitem{wang22-uformer}
Wang, Z., Cun, X., Bao, J., Zhou, W., Liu, J., Li, H.: Uformer: A general u-shaped transformer for image restoration. In: CVPR. pp. 17683--17693 (June 2022)

\bibitem{wang04}
Wang, Z., Bovik, A., Sheikh, H., Simoncelli, E.: Image quality assessment: from error visibility to structural similarity. IEEE TIP  \textbf{13}(4),  600--612 (2004)

\bibitem{xie22}
Xie, Y., Takikawa, T., Saito, S., Litany, O., Yan, S., Khan, N., Tombari, F., Tompkin, J., Sitzmann, V., Sridhar, S.: Neural fields in visual computing and beyond. Comput. Graph. Forum  (2022)

\bibitem{xu23}
Xu, L., Agrawal, V., Laney, W., Garcia, T., Bansal, A., Kim, C., Rota~Bulò, S., Porzi, L., Kontschieder, P., Božič, A., Lin, D., Zollhöfer, M., Richardt, C.: {VR-NeRF}: High-fidelity virtualized walkable spaces. In: SIGGRAPH Asia Conference Proceedings (2023)

\bibitem{yan19}
Yan, Q., Gong, D., Shi, Q., Hengel, A.v.d., Shen, C., Reid, I., Zhang, Y.: Attention-guided network for ghost-free high dynamic range imaging. In: CVPR. pp. 1751--1760 (2019)

\bibitem{yu21-2}
Yu, A., Li, R., Tancik, M., Li, H., Ng, R., Kanazawa, A.: {PlenOctrees} for real-time rendering of neural radiance fields. In: ICCV (2021)

\bibitem{zhang20}
Zhang, K., Riegler, G., Snavely, N., Koltun, V.: Nerf++: Analyzing and improving neural radiance fields. arXiv:2010.07492v2  (2020)

\bibitem{zhang18}
Zhang, R., Isola, P., Efros, A.A., Shechtman, E., Wang, O.: The unreasonable effectiveness of deep features as a perceptual metric. In: CVPR. pp. 586--595 (2018)

\bibitem{zhou23pami}
Zhou, K., Li, W., Jiang, N., Han, X., Lu, J.: From nerflix to nerflix++: A general nerf-agnostic restorer paradigm. IEEE TPAMI pp. 1--17 (2023)

\bibitem{zhou23}
Zhou, K., Li, W., Wang, Y., Hu, T., Jiang, N., Han, X., Lu, J.: Nerflix: High-quality neural view synthesis by learning a degradation-driven inter-viewpoint mixer. In: CVPR. pp. 12363--12374 (2023)

\end{thebibliography}


\begin{thebibliography}{10}
\providecommand{\url}[1]{\texttt{#1}}
\providecommand{\urlprefix}{URL }
\providecommand{\doi}[1]{https://doi.org/#1}

\bibitem{barron22}
Barron, J.T., Mildenhall, B., Verbin, D., Srinivasan, P.P., Hedman, P.: Mip-nerf 360: Unbounded anti-aliased neural radiance fields. In: CVPR (2022)

\bibitem{chen22}
Chen, A., Xu, Z., Geiger, A., Yu, J., Su, H.: Tensorf: Tensorial radiance fields. In: ECCV (2022)

\bibitem{he15}
He, K., Zhang, X., Ren, S., Sun, J.: Delving deep into rectifiers: Surpassing human-level performance on imagenet classification. In: ICCV. pp. 1026--1034 (2015)

\bibitem{jiang23}
Jiang, Y., Hedman, P., Mildenhall, B., Xu, D., Barron, J.T., Wang, Z., Xue, T.: Alignerf: High-fidelity neural radiance fields via alignment-aware training. In: CVPR. pp. 46--55 (2023)

\bibitem{mildenhall19}
Mildenhall, B., Srinivasan, P.P., Ortiz-Cayon, R., Kalantari, N.K., Ramamoorthi, R., Ng, R., Kar, A.: Local light field fusion: Practical view synthesis with prescriptive sampling guidelines. ACM TOG  (2019)

\bibitem{mildenhall20}
Mildenhall, B., Srinivasan, P.P., Tancik, M., Barron, J.T., Ramamoorthi, R., Ng, R.: Nerf: Representing scenes as neural radiance fields for view synthesis. In: ECCV (2020)

\bibitem{pytorch}
Paszke, A., Gross, S., Massa, F., Lerer, A., Bradbury, J., Chanan, G., Killeen, T., Lin, Z., Gimelshein, N., Antiga, L., Desmaison, A., Kopf, A., Yang, E., DeVito, Z., Raison, M., Tejani, A., Chilamkurthy, S., Steiner, B., Fang, L., Bai, J., Chintala, S.: {PyTorch: An Imperative Style, High-Performance Deep Learning Library}. In: Wallach, H., Larochelle, H., Beygelzimer, A., d'Alché Buc, F., Fox, E., Garnett, R. (eds.) NeurIPS. pp. 8024--8035. Curran Associates, Inc. (2019)

\bibitem{eriba2019kornia}
Riba, E., Mishkin, D., Ponsa, D., Rublee, E., Bradski, G.: Kornia: an open source differentiable computer vision library for pytorch. In: Winter Conference on Applications of Computer Vision (2020)

\bibitem{tancik23}
Tancik, M., Weber, E., Ng, E., Li, R., Yi, B., Kerr, J., Wang, T., Kristoffersen, A., Austin, J., Salahi, K., Ahuja, A., McAllister, D., Kanazawa, A.: Nerfstudio: A modular framework for neural radiance field development. In: ACM SIGGRAPH 2023 Conference Proceedings. SIGGRAPH '23 (2023)

\bibitem{wang22-uformer}
Wang, Z., Cun, X., Bao, J., Zhou, W., Liu, J., Li, H.: Uformer: A general u-shaped transformer for image restoration. In: CVPR. pp. 17683--17693 (June 2022)

\bibitem{xue19}
Xue, T., Chen, B., Wu, J., Wei, D., Freeman, W.: Video enhancement with task-oriented flow. IJCV  \textbf{127} (08 2019)

\bibitem{zhou23pami}
Zhou, K., Li, W., Jiang, N., Han, X., Lu, J.: From nerflix to nerflix++: A general nerf-agnostic restorer paradigm. IEEE TPAMI pp. 1--17 (2023)

\bibitem{zhou23}
Zhou, K., Li, W., Wang, Y., Hu, T., Jiang, N., Han, X., Lu, J.: Nerflix: High-quality neural view synthesis by learning a degradation-driven inter-viewpoint mixer. In: CVPR. pp. 12363--12374 (2023)

\end{thebibliography}









\end{document}


\title{RoGUENeRF: A Robust Geometry-Consistent Universal Enhancer for NeRF \\
Supplementary Material}

\newcommand{\ours}{RoGUENeRF}
\newcommand{\cellbest}{\cellcolor{red!20}}
\newcommand{\cellsecond}{\cellcolor{orange!20}}
\newcommand{\boxtick}{\makebox[0pt][l]{$\square$}\raisebox{.15ex}{\hspace{0.1em}$\checkmark$}}

\author{Sibi Catley-Chandar\inst{1,2} \and
Richard Shaw\inst{1}\and \\
Gregory Slabaugh\inst{2}\and 
Eduardo P\'erez-Pellitero\inst{1}}

\titlerunning{RoGUENeRF - Supplementary Material}

\authorrunning{S.~Catley-Chandar et al.}


\institute{Huawei Noah's Ark Lab\and 
Queen Mary University of London}

\maketitle



\begin{figure}[ht]
\centering
\includegraphics[width=1.0\textwidth]{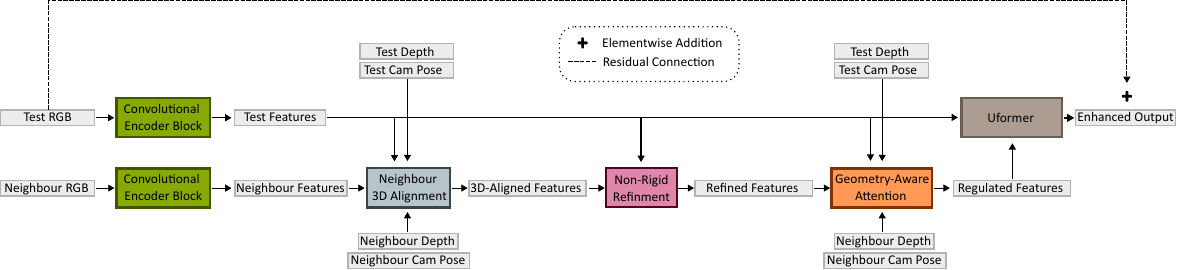}
\caption{An overview of our entire pipeline. We present implementation details for each component of our pipeline in Section \ref{sec:implementation}.}
\label{fig:pipeline}
\end{figure}

We present further detailed qualitative and quantitative evaluation of our method in Sections \ref{sec:video} and \ref{sec:per-scene}. We evaluate the amount of training data required to train our model compared to the state-of-the-art in Section \ref{sec:data} and compare inference speeds in Section \ref{sec:inference}. We present full implementation details of each component of our method in Section \ref{sec:implementation}.

\section{Video Results}
\label{sec:video}

We present video results of our proposed enhancer compared to NeRFLiX \cite{zhou23}, MipNeRF360 \cite{barron22}, Nerfacto \cite{tancik23} and NeRF \cite{mildenhall20} on our project page: \url{https://sib1.github.io/projects/roguenerf/}. We demonstrate noticeable improvements over the baseline NeRF models and NeRFLiX, whilst retaining geometric consistency and restoring high-frequency textures. We encourage viewing the video results to fully appreciate the improvement in quality provided by our method.

 \begin{figure}[!t]
\centering
\includegraphics[width=0.9\textwidth]{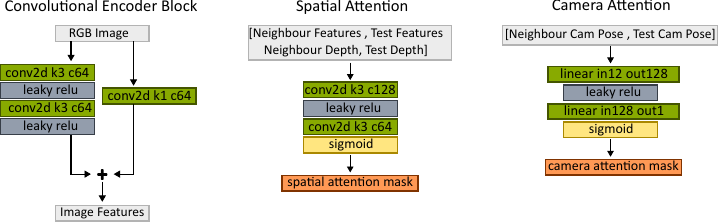}
\caption{We present detailed architectures of our convolutional encoder and geometry-aware attention modules. k indicates kernel size, c indicates output channel size, in/out indicate input/output dimension size.}
\label{fig:attention}
\end{figure}

\section{Real World Data Efficiency}
\label{sec:data}

Real world NeRF datasets are typically small, sometimes only containing tens of images per scene. Exisiting NeRF enhancers like NeRFLiX \cite{zhou23} struggle to leverage such small datasets, requiring several orders of magnitude more training data. In contrast, our novel geometry-consistent alignment and fusion mechanisms allow our enhancer to be pre-trained and fine-tuned on relatively small datasets. We compare the results of training NeRFLiX on different dataset sizes with our method in Table \ref{table:train-nerflix}. We show that NeRFLiX requires several orders of magnitude more simulated training data (Vimeo90K \cite{xue19}) than our method to achieve any improvement at all over the baseline NeRF model, TensoRF \cite{chen22}. When trained on only the real world LLFF dataset, NeRFLiX actually degrades the performance of TensoRF by 0.6dB. In contrast, our method can be pre-trained and fine-tuned on small datasets to achieve large improvements in performance.

  \begin{table}[h]
\centering
\caption{We show that our method is substantially more data efficient than NeRFLiX as it can be pre-trained and fine-tuned on very small datasets. NeRFLiX is unable to learn an improvement over the baseline using only a small real world dataset. $^\dagger$Results as reported by authors.}
\begin{adjustbox}{width=0.9\textwidth}
\begin{tabular}{l l c c } \\ \toprule
  Model &\hspace{0.5cm}Training Data & \# Training Frames & PSNR\\ \midrule
  TensoRF &\hspace{0.5cm} LLFF & 262 &\hspace{-1.3cm} 26.88  \\ 
    TensoRF + NeRFLiX$^\dagger$  &\hspace{0.5cm} LLFF & 262 &  26.28 ($\downarrow$ 0.60) \\
    TensoRF + NeRFLiX$^\dagger$  &\hspace{0.5cm} LLFF + Vimeo90K-10\%&   45,490 & 26.71 ($\downarrow$ 0.17) \\
    TensoRF + NeRFLiX$^\dagger$  &\hspace{0.5cm} LLFF + Vimeo90K-50\%&  226,404 & 27.08 ($\uparrow$ 0.30)\\
      TensoRF + NeRFLiX  &\hspace{0.5cm} LLFF + Vimeo90K-100\%&  452,546 &\cellsecond 27.38 ($\uparrow$ 0.50)  \\
   TensoRF + Ours  &\hspace{0.5cm} LLFF & 262 & \cellbest 27.58 ($\uparrow$ 0.70)   \\ 
\bottomrule
\end{tabular}
\end{adjustbox} 
\label{table:train-nerflix}
 \end{table}

\section{Inference Speed}
\label{sec:inference}

We compare the inference speed of NeRFLiX and NeRFLiX++ \cite{zhou23pami} with our method for different image resolutions in Table \ref{table:inference-speed}. We measure inference speed on a single NVidia V100 GPU at full float32 precision using the Pytorch event time measurement function \cite{pytorch}. For NeRFLiX++, we report inference times directly from the original work. We note that for all quantitative and qualitative comparisons in this work, NeRFLiX performs test-time augmentation (TTA) by running 4$\times$ forward passes during inference with horizontally and vertically flipped inputs and averaging the result. We perform a single inference pass and do not use TTA. We show that \ours \ is approximately 13$\times$ faster than NeRFLiX with TTA, 3$\times$ faster than NeRFLiX without TTA, and 3$\times$ slower than NeRFLiX++. 

  \begin{table}[h]
\centering
\caption{ We present quantitative results from using different datasets for training. TTA = Test-Time Augmentation. NeRFLiX results reported in this work use TTA.}
\begin{adjustbox}{width=0.7\textwidth}
\begin{tabular}{l c c c } \\ \toprule
  Model & TTA &\hspace{0.5cm}Input Resolution &\hspace{0.5cm}Inference Time (ms)\\  \midrule
       NeRFLiX & Yes & 512x512 &  4208 \\
       NeRFLiX & No & 512x512 &  1050 \\
    Ours  &  No & 512x512& \cellsecond337  \\ 
    NeRFLiX++  &  No & 512x512& \cellbest 109 \\ \midrule
        NeRFLiX& Yes & 1024x1024 &  18385 \\
        NeRFLiX & No & 1024x1024 &  4579 \\
   Ours & No & 1024x1024 &   \cellsecond 1346  \\ 
    NeRFLiX++  &  No & 1024x1024& \cellbest 433 \\
\bottomrule
\end{tabular}
\end{adjustbox} 
\label{table:inference-speed}
 \end{table}

\section{Implementation Details}
\label{sec:implementation}

We present our entire pipeline in Figure \ref{fig:pipeline}. For each of the learnable components in our pipeline, we present detailed module architectures in Figures \ref{fig:attention} and \ref{fig:flow}. We refer to a standard 2d convolutional layer as $conv2d$, and indicate the kernel size with $k$, channel output size with $c$, stride with $s$ and input/output dimensions for linear layers with $in$/$out$. We use zero padding and He initialization \cite{he15} for learnable weights. We use leaky relu with a slope of 0.01. 

\subsection{3D Alignment}

We first extract features using the convolutional encoder block shown in Figure \ref{fig:attention}. We use the Kornia \cite{eriba2019kornia} geometry remap function with bilinear interpolation and zero padding to perform the 3D alignment of neighbouring training images to a novel test view. The inputs to the function are formulated as described in the main manuscript.

\subsection{Iterative Refinement Network}

\begin{figure}[h]
\centering
\includegraphics[width=0.45\textwidth]{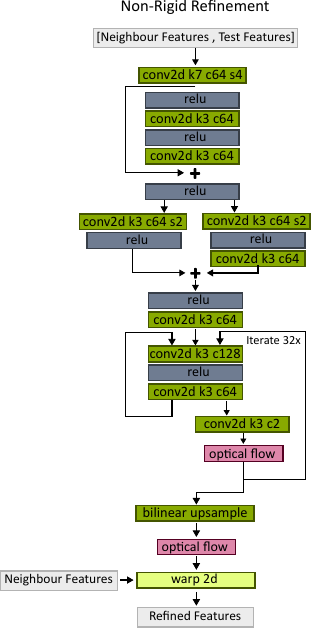}
\caption{We present the implementation details of our iterative flow network for non-rigid refinement. The optical flow estimation is performed at 8$\times$ downsampled resolution and we apply bilinear upsampling to recover a full resolution flow field.}
\label{fig:flow}
\end{figure}

We present the implementation details of our iterative refinement network in Figure \ref{fig:flow}. We perform our optical flow estimation at 8$\times$ downsampled resolution and apply bilinear upsampling to the estimated flow field to get back to full resolution. Our flow warping uses the PyTorch grid sample function with bilinear interpolation.

\subsection{Geometry-Aware Attention}

We present the architecture of our spatial attention and camera attention modules in Figure \ref{fig:attention}. We learn a bias for each convolutional and linear layer. Camera rotations are first converted to Euler angles before being processed by the attention module. 

\subsection{Uformer}

We use the official implementation of the Uformer \cite{wang22-uformer}. We modify the original architecture to accept 64 channel image features as input instead of 3 channel RGB images. Otherwise we keep the architecture identical to the original formulation.

\section{Per-Scene Results}
\label{sec:per-scene}

We present extended quantitative results in Tables \ref{table:per-scene-360} and \ref{table:per-scene-llff} where we report results for each scene in the 360v2 \cite{barron22} and LLFF \cite{mildenhall19} datasets. We show that \ours \ consistently outperforms the baseline NeRF models, NeRFLiX and AligNeRF \cite{jiang23} on the majority of scenes across all metrics.

 \begin{table*}[h]
\centering
\caption{Per-Scene results on the 360v2 dataset. All scenes are evaluated at 4$\times$ downsampled resolution. Red / orange highlights indicate best / second method respectively. $^\dagger$Results as reported by authors.}
\begin{subtable}[h]{1.0\textwidth}
\caption{PSNR}
\begin{adjustbox}{width=1.0\textwidth}
\begin{tabular}{l c c c c c c c c c } \\ \toprule
  Model & Bicycle & Bonsai & Counter &Garden& Kitchen & Room &Stump & Flowers & Treehill  \\ \midrule
    ZipNeRF  &  25.88 & 35.65 & 29.5 & \cellsecond28.24 & 33.28 & 34.03 & 27.32 & 22.32 & 23.91\\
    ZipNeRF + NeRFLiX &  \cellsecond25.9 & \cellsecond36.22 &  \cellbest30.06 & 27.33 & \cellsecond33.42 & \cellsecond34.16 & \cellsecond27.44 & \cellsecond22.42 & \cellbest24.05\\
    ZipNeRF + Ours   &  \cellbest26.2 & \cellbest36.4 & \cellsecond29.84 & \cellbest28.57 & \cellbest333.67 & \cellbest34.32 & \cellbest27.55 & \cellbest22.55 & \cellsecond24.02\\ \hdashline
    
    MipNeRF360  & 24.31 & 33.99 & 29.83 & 27.00 & 32.91 & 32.53 & 26.36 & \cellsecond21.68 & \cellsecond25.72\\
    MipNeRF360 + NeRFLiX & 24.22 & \cellsecond34.98 & \cellsecond30.49 & 26.40 & \cellsecond33.31 & \cellsecond32.91 & 26.49 & 21.64 & 25.56\\
    MipNeRF360 + AligNeRF$^\dagger$ & \cellsecond24.75 & - & - & \cellsecond27.07 & - & - & \cellsecond26.69& 21.61 & -\\
    MipNeRF360 + Ours   & \cellbest24.81 & \cellbest35.17 & \cellbest30.53 & \cellbest27.49 & \cellbest34.0 & \cellbest33.06 & \cellbest26.82 & \cellbest22.05 & \cellbest26.07\\\hdashline
    
    Nerfacto &23.16 & 29.88 & 27.14 & 25.28 & 30.09 & 30.77 & 25.31 & 20.71 & 22.69\\
    Nerfacto + NeRFLiX & \cellsecond23.26 & \cellsecond31.92 &\cellsecond 28.71 &\cellsecond 25.62 &\cellsecond 31.86 &\cellsecond 31.78 &\cellsecond 25.59 & \cellsecond20.79 & \cellsecond22.75\\
    Nerfacto + Ours  & \cellbest23.79 & \cellbest32.79 & \cellbest28.71 & \cellbest26.46 & \cellbest32.57 & \cellbest32.56 &\cellbest 25.86 &\cellbest 21.26 &\cellbest 23.08\\\midrule
    \vspace{0.5cm}
\end{tabular}
\end{adjustbox}

\end{subtable}

\begin{subtable}[h]{1.0\textwidth}
\centering
\caption{SSIM}
\begin{adjustbox}{width=1.0\textwidth}
\begin{tabular}{l c c c c c c c c c} \\ \toprule
  Model & Bicycle & Bonsai & Counter &Garden& Kitchen & Room &Stump  & Flowers & Treehill \\ \midrule
    ZipNeRF  & \cellsecond 0.7729 & 0.968 & 0.9215 & \cellsecond 0.8631 & \cellsecond 0.9522 & 0.953 & 0.7881 &\cellsecond0.6366 &\cellsecond 0.6749\\
    ZipNeRF + NeRFLiX &  0.7611 &\cellsecond 0.9714 & \cellbest0.9337 & 0.8245 & 0.9521 & \cellsecond 0.9536 & \cellsecond 0.7898 & 0.6257 & 0.6731\\
    ZipNeRF + Ours  & \cellbest 0.7877 & \cellbest0.972 & \cellsecond 0.9313 & \cellbest0.8732 & \cellbest0.956 & \cellbest0.9569 & \cellbest0.8002 & \cellbest0.6567 & \cellbest0.6842\\\hdashline
    
    MipNeRF360  &  0.6798 & 0.9554 & 0.9082 & 0.8130 & 0.9425 & 0.9374 & 0.7428 & 0.5786 &\cellsecond 0.6876\\
    MipNeRF360 + NeRFLiX & 0.6696 & \cellsecond 0.9656 & \cellsecond 0.9279 & 0.7843 & \cellsecond 0.9499 &\cellsecond 0.9442 & 0.7467 & 0.5666 & 0.6776\\
    MipNeRF360 + AligNeRF$^\dagger$& \cellsecond 0.7052 & - & - & \cellsecond 0.8250 & - & - & \cellsecond 0.7650 & \cellsecond 0.5880 & - \\
    MipNeRF360 + Ours   & \cellbest 0.7225 & \cellbest 0.9669 & \cellbest 0.9301 & \cellbest 0.8428 & \cellbest 0.9565 & \cellbest 0.9508 & \cellbest 0.7716 & \cellbest 0.6174 & \cellbest 0.7136\\\hdashline
    
    Nerfacto  & 0.5286 & 0.9082 & 0.8299 & 0.7151 & 0.9005 & 0.9001 & 0.6628 & 0.474 & 0.5216\\
    Nerfacto + NeRFLiX &\cellsecond   0.5334 &\cellsecond  0.9448 & \cellsecond 0.8954 & \cellsecond 0.7386 &\cellsecond  0.9378 & \cellsecond 0.926 &\cellsecond  0.679 & \cellsecond 0.4768 & \cellsecond 0.5369\\
    Nerfacto + Ours & \cellbest0.5848 & \cellbest0.9496 & \cellbest0.8985 &\cellbest 0.7976 & \cellbest0.9413 & \cellbest0.9408 &\cellbest 0.7063 &\cellbest 0.5398 &\cellbest 0.571 \\\midrule
    \vspace{0.5cm}
\end{tabular}
\end{adjustbox}

\end{subtable}

 \begin{subtable}[h]{1.0\textwidth} 
\centering
\caption{LPIPS}
\begin{adjustbox}{width=1.0\textwidth}
\begin{tabular}{l c c c c c c c c c} \\ \toprule
  Model & Bicycle & Bonsai & Counter &Garden& Kitchen & Room &Stump   & Flowers & Treehill\\ \midrule
    ZipNeRF  & \cellsecond 0.2299 & \cellbest0.0886 & \cellsecond 0.1227 & \cellsecond 0.1283 & \cellbest0.0726 & \cellbest0.1266 & \cellsecond 0.2410 & \cellsecond 0.3107 & \cellsecond 0.2811\\
    ZipNeRF + NeRFLiX & 0.2747 & 0.101 & 0.1230 & 0.1858 & 0.0831 & 0.1485 & 0.2605 & 0.3348 & 0.3292\\
    ZipNeRF + Ours   &   \cellbest0.2096 & \cellsecond 0.0924 & \cellbest0.1149 & \cellbest0.1189 & \cellsecond 0.0756 & \cellsecond 0.1290 & \cellbest0.2209 & \cellbest0.263 &\cellbest 0.2715\\\hdashline
    
    MipNeRF360  &  \cellsecond 0.3406 & \cellsecond 0.1065 & 0.1504 &\cellsecond  0.1911 & 0.0879 &\cellsecond  0.1505 & \cellsecond 0.3072 &\cellsecond  0.3763 &\cellsecond  0.3568\\
    MipNeRF360 + NeRFLiX &  0.3696 & 0.1098 & \cellsecond 0.1332 & 0.2248 &\cellsecond  0.0853 & 0.1615 & 0.3172 & 0.4032 & 0.3919\\
    MipNeRF360 + Ours  &  \cellbest0.2892 & \cellbest0.1055 & \cellbest0.1227 & \cellbest0.1516 &\cellbest 0.0761 & \cellbest0.1393 & \cellbest0.2653 & \cellbest0.3206 & \cellbest0.3177\\\hdashline
    
    Nerfacto  &  \cellsecond 0.4988 & 0.1749 & 0.2524 & 0.2956 & 0.1401 & 0.2138 & 0.3852 & \cellsecond 0.4646 & \cellsecond 0.5135\\
    Nerfacto + NeRFLiX & 0.5067 & \cellsecond  0.1362 &\cellsecond  0.1703 &\cellsecond  0.2696 & \cellsecond 0.0989 & \cellsecond 0.1871 & \cellsecond 0.3707 & 0.4742 & 0.5258\\
    Nerfacto + Ours &\cellbest 0.4436 &\cellbest 0.1298 & \cellbest0.1598 &\cellbest 0.2077 & \cellbest0.0944 & \cellbest0.1518 & \cellbest0.3394 &\cellbest 0.4044 & \cellbest0.4723\\\midrule
\end{tabular}
\end{adjustbox}

\end{subtable}

\label{table:per-scene-360}
 \end{table*}

 \begin{table*}[h]
\centering
\caption{Per-Scene results on the LLFF dataset.  All scenes are evaluated at 4$\times$ downsampled resolution. Red / orange highlights indicate best / second method respectively.}
\begin{subtable}[h]{0.85\textwidth}
\caption{PSNR}
\begin{adjustbox}{width=1.0\textwidth}
\begin{tabular}{l c c c c c c c c} \\ \toprule
  Model & Fern & Flower & Fortress  & Horns & Leaves & Orchids & Room & T-Rex  \\ \midrule
    NeRF  & 24.90 & 27.76 & 31.43 & 27.55 & 21.06 & 20.24 & 32.62 & 27.03\\
    NeRF + NeRFLiX & \cellbest{25.48} & \cellsecond28.42 & \cellsecond31.89 & \cellsecond28.57 & \cellsecond21.24 & \cellbest{20.41} & \cellsecond33.44 & \cellsecond27.90\\
    NeRF + Ours   &  \cellsecond25.44 & \cellbest{28.43} & \cellbest{32.55} & \cellbest{29.86} & \cellbest{22.02} & \cellsecond20.37 & \cellbest{33.95} & \cellbest{28.71}\\\hdashline
    
    TensoRF  &  24.37 & 28.87 & 31.41 & 28.98 & 21.46 & 19.45 & 32.77 & 27.70\\
    TensoRF + NeRFLiX & \cellbest{25.90} & \cellsecond29.21 & \cellsecond31.72 & \cellsecond29.20 & \cellsecond21.50 & \cellbest{20.16} & \cellsecond33.58 & \cellsecond27.76\\
    TensoRF + Ours   &  \cellsecond24.76 & \cellbest{29.27} & \cellbest{32.05} & \cellbest{30.51} & \cellbest{21.85} & \cellsecond19.62 & \cellbest{33.87} & \cellbest{28.71}\\\midrule
    \vspace{0.5cm}
\end{tabular}
\end{adjustbox}
\end{subtable}

\begin{subtable}[h]{0.85\textwidth}
\centering
\caption{SSIM}
\begin{adjustbox}{width=1.0\textwidth}
\begin{tabular}{l c c c c c c c c} \\ \toprule
  Model & Fern & Flower & Fortress  & Horns & Leaves & Orchids & Room &T-Rex  \\ \midrule
    NeRF  & 0.7917 & 0.8393 & 0.8863 & 0.8342 & 0.6995 & 0.6466 & 0.9511 & 0.8873\\
    NeRF + NeRFLiX &  \cellsecond 0.8382 & \cellsecond0.8752 &\cellsecond 0.9045 & \cellsecond0.8954 &\cellsecond 0.7543 & \cellsecond0.6912 &\cellsecond 0.9638 & \cellsecond0.9188\\
    NeRF + Ours   & \cellbest{0.8470} & \cellbest{0.8791} & \cellbest{0.9183} & \cellbest{0.9239} & \cellbest{0.7945} & \cellbest{0.7039} & \cellbest{0.9676} & \cellbest{0.9356} \\\hdashline
    
    TensoRF  & 0.7875 & 0.8791 & 0.9001 & 0.9006 & 0.7709 & 0.6363 & 0.9578 & 0.9136\\
    TensoRF + NeRFLiX &  \cellbest{0.8533} & \cellsecond0.8916 &\cellsecond 0.9072 &\cellsecond 0.9161 & \cellsecond0.7795 & \cellbest{0.6842} & \cellsecond0.9658 &\cellsecond 0.9240\\
    TensoRF + Ours   &  \cellsecond0.8184 & \cellbest{0.8935} & \cellbest{0.9123} & \cellbest{0.9364} & \cellbest{0.8095} & \cellsecond0.6607 & \cellbest{0.9674} & \cellbest{0.9376}\\\midrule
    \vspace{0.5cm}
\end{tabular}
\end{adjustbox}
\end{subtable}

 \begin{subtable}[h]{0.85\textwidth} 
\centering
\caption{LPIPS}
\begin{adjustbox}{width=1.0\textwidth}
\begin{tabular}{l c c c c c c c c} \\ \toprule
  Model & Fern & Flower & Fortress  & Horns & Leaves & Orchids & Room &T-Rex  \\ \midrule
    NeRF  &0.2747 & 0.2036 & 0.1546 & 0.2590 & 0.3038 & 0.3083 & 0.1639 & 0.2430\\
    NeRF + NeRFLiX & \cellsecond0.1902 & \cellsecond0.1490 & \cellsecond0.1486 & \cellsecond0.1615 & \cellsecond0.1779 & \cellsecond0.2224 & \cellsecond0.1248 & \cellsecond0.1818\\
    NeRF + Ours   & \cellbest{0.1799} & \cellbest{0.1377} & \cellbest{0.1128} & \cellbest{0.1295} & \cellbest{0.1469} & \cellbest{0.2043} & \cellbest{0.1198} & \cellbest{0.1651}\\\hdashline
    
    TensoRF  & 0.2373 & 0.1417 & 0.1266 & 0.1522 & 0.2067 & 0.2657 & 0.1402 & 0.1927\\
    TensoRF + NeRFLiX & \cellbest{0.1693} & \cellsecond0.1304 & \cellsecond0.1352 & \cellsecond0.1288 & \cellsecond0.1510 & \cellbest{0.2086} &\cellbest{ 0.1159} & \cellsecond0.1720\\
    TensoRF + Ours   &  \cellsecond0.1887 & \cellbest{0.1259} & \cellbest{0.1206} & \cellbest{0.1131} &\cellbest{0.1406} & \cellsecond0.2220 & \cellsecond0.1179 & \cellbest{0.1660}\\\midrule
\end{tabular}
\end{adjustbox}
\end{subtable}

\label{table:per-scene-llff}
 \end{table*}

\clearpage

\bibliographystyle{splncs04}
\bibliography{main}